\documentclass[runningheads]{llncs}
\usepackage{graphicx}
\usepackage{amssymb}
\usepackage{amsmath}
\usepackage{multirow}
\usepackage{hhline}
\usepackage[table,x11names]{xcolor}
\usepackage{hyperref}
\hypersetup{colorlinks=true,citecolor=teal}
\usepackage{xcolor}
\usepackage[draft]{pgf}

\newcommand*\samethanks[1][\value{footnote}]{\footnotemark[#1]}
\begin{document}
\title{Synthetic Augmentation with Large-scale Unconditional Pre-training} 

\author{Jiarong Ye\inst{1}\thanks{These authors contributed equally to this work.} \and Haomiao Ni\inst{1}\samethanks \and Peng Jin\inst{1} \and Sharon X. Huang\inst{1} \and Yuan Xue\inst{2,3}}

\authorrunning{Ye et al.}
\institute{The Pennsylvania State University, University Park, Pennsylvania, USA \and Johns Hopkins University, Baltimore, Maryland, USA 
\and
The Ohio State University, Columbus, Ohio, USA\\
\email{Yuan.Xue@osumc.edu}
}
%
%
%
\maketitle              
\begin{abstract}
Deep learning based medical image recognition systems often require a substantial amount of training data with expert annotations, which can be expensive and time-consuming to obtain. Recently, synthetic augmentation techniques have been proposed to mitigate the issue by generating realistic images conditioned on class labels. However, the effectiveness of these methods heavily depends on the representation capability of the trained generative model, which cannot be guaranteed without sufficient labeled training data. To further reduce the dependency on annotated data, we propose a synthetic augmentation method called HistoDiffusion, which can be pre-trained on large-scale unlabeled datasets and later applied to a small-scale labeled dataset for augmented training. In particular, we train a latent diffusion model (LDM) on diverse unlabeled datasets to learn common features and generate realistic images without conditional inputs. Then, we fine-tune the model with classifier guidance in latent space on an unseen labeled dataset so that the model can synthesize images of specific categories. Additionally, we adopt a selective mechanism to only add synthetic samples with high confidence of matching to target labels. We evaluate our proposed method by pre-training on three histopathology datasets and testing on a histopathology dataset of colorectal cancer (CRC) excluded from the pre-training datasets. With HistoDiffusion augmentation, the classification accuracy of a backbone classifier is remarkably improved by 6.4\% using a small set of the original labels. Our code is available at \href{https://github.com/karenyyy/HistoDiffAug}{\color{blue}{https://github.com/karenyyy/HistoDiffAug}}.


\end{abstract}
\section{Introduction}

The recent advancements in medical image recognition systems have greatly benefited from deep learning techniques~\cite{ker2017deep,shen2017deep}. Large-scale well-annotated datasets are one of the key components for training deep learning models to achieve satisfactory results~\cite{deng2009imagenet,lin2014microsoft}. However, unlike natural images in computer vision, the number of medical images with expert annotations is often limited by the high labeling cost and privacy concerns. To overcome this challenge, a natural choice is to employ data augmentation to increase the number of training samples. Although conventional augmentation techniques~\cite{perez2017effectiveness} such as flipping and cropping can be directly applied to medical images, they merely improve the diversity of datasets, thus leading to marginal performance gains~\cite{chen2022generative}. Another group of studies employ conditional generative adversarial networks (cGANs)~\cite{goodfellow2020generative} to synthesize visually appealing medical images that closely resemble those in the original datasets~\cite{xue2021selective,xue2019synthetic}. While existing works have proven effective in improving the performance of downstream models to some extent, a sufficient amount of labeled data is still required to adequately train models to generate decent-quality images. More recently, diffusion models have become popular for natural image generation due to their impressive results and training stability~\cite{dhariwal2021diffusion,ho2020denoising,song2020denoising}. A few studies have also demonstrated the potential of diffusion models for medical image synthesis~\cite{moghadam2023morphology,pinaya2022brain}.  

\begin{figure}[t]
\centering
\includegraphics[width=0.8\textwidth]{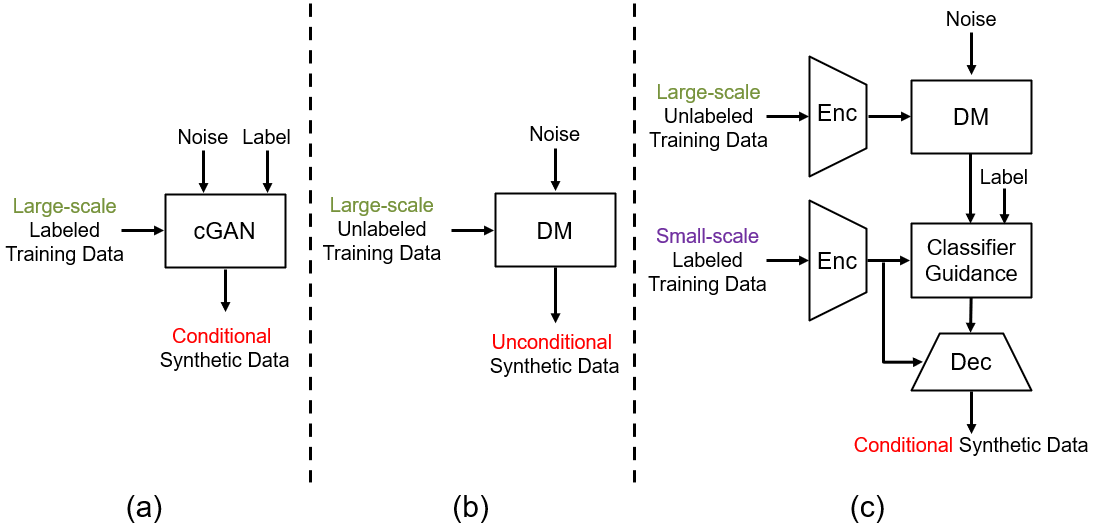}
\caption{Comparison between different deep generative models for synthetic augmentation. (a) cGAN-based method which requires relatively large-scale annotated training data; (b) Diffusion model (DM) which cannot take conditional input; (c) Our proposed HistoDiffusion model that can be pretrained on large-scale unannotated data and later applied to \textit{unseen} small-scale annotated data for augmentation.} 
\label{fig:teaser}
\vspace{-3mm}
\end{figure}

Although annotated data is typically hard to acquire for medical images, unannotated data is often more accessible. To mitigate the issue existed in current cGAN-based synthetic augmentation methods~\cite{frid2018gan,xue2021selective,xue2019synthetic,ye2020synthetic}, in this work, we propose to leverage the diffusion model with unlabeled pre-training to reduce the dependency on the amount of labeled data (see comparisons in Fig.~\ref{fig:teaser}). We propose a novel synthetic augmentation method, named HistoDiffusion, which can be pre-trained on large-scale unannotated datasets and adapted to small-scale annotated datasets for augmented training. Specifically, we first employ a latent diffusion model (LDM) and train it on a collection of unlabeled datasets from multiple sources. This large-scale pre-training enables the model to learn common yet diverse image characteristics and generate realistic medical images. Second, given a small labeled dataset that does not exist in the pre-training datasets, the decoder of the LDM is fine-tuned using annotations to adapt to the domain shift. Synthetic images are then generated with classifier guidance~\cite{dhariwal2021diffusion} in the latent space.  Following the prior work~\cite{xue2021selective}, we select generated images based on the confidence of target labels and feature similarity to real labeled images. We evaluate our proposed method on a histopathology image dataset of colorectal cancer (CRC). Experiment results show that when presented with limited annotations, the classifier trained with our augmentation method outperforms the ones trained with the prior cGAN-based methods. Our experimental results show that once HistoDiffusion is well pre-trained using large datasets, it can be applied to any future incoming small dataset with minimal fine-tuning and may substantially improve the flexibility and efficacy of synthetic augmentation.  




\section{Methodology}
\begin{figure}[t]
\centering
\includegraphics[width=\textwidth]{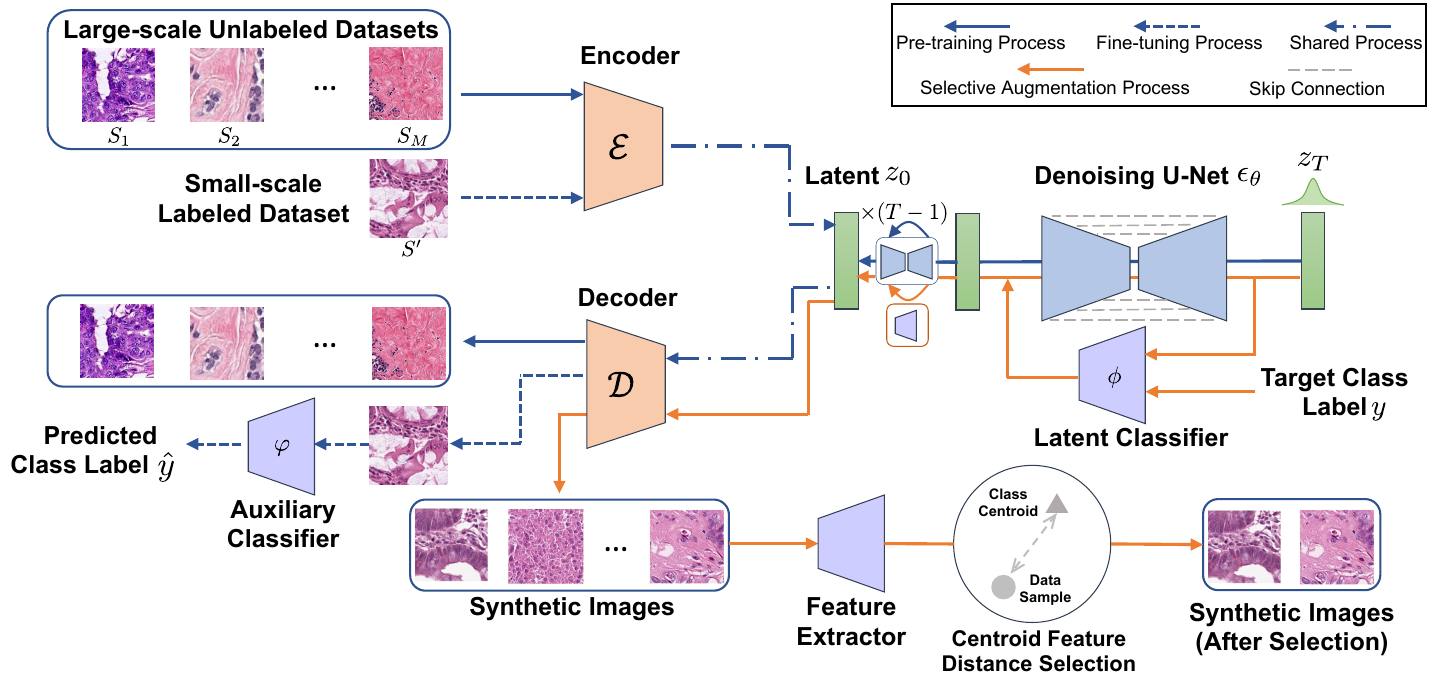}
\caption{The architecture of our proposed HistoDiffusion, which consists of a pre-training process (blue solid lines), a fine-tuning process (blue dashed lines), and a selective augmentation process (orange lines). During pre-training, a latent autoencoder (LAE) and a diffusion model (DM) are trained on large-scale unlabeled datasets for unconditional image synthesis. HistoDiffusion is then fine-tuned on a small-scale dataset for conditional image synthesis under the guidance of a trained latent classifier. During selective augmentation, given a target class label, the synthetic images generated by the fine-tuned model are selected and added to the training set based on their distances to the class centroids in the feature space.}
\label{fig:arch}
\vspace{-2mm}
\end{figure}

Fig.~\ref{fig:arch} illustrates the overall architecture of our proposed method. First, we train an LDM on a large-scale set of unlabeled datasets collected from multiple sources. We then fine-tune the decoder of this pretrained LDM on a small labeled dataset. To enable conditional image synthesis, we also train a latent classifier on the same labeled dataset to guide the diffusion model in LDM. Once the classifier is trained, we apply the fine-tuned LDM to generate a pool of candidate images conditioned on the target class labels. These candidate images are then passed through the image selection module to filter out any low-quality results. Finally, we can train downstream classification models on the expanded training data, which includes the selected images, and then use them to perform inference on test data. 
In this section, we will first introduce the background of diffusion models and then present details about the HistoDiffusion model.

\subsection{Diffusion Models}
Diffusion models (DM) \cite{ho2020denoising,sohl2015deep,song2019generative} are probabilistic models that are designed to
learn a data distribution. 
Given a sample from the data distribution $z_0\sim q(z_0)$, the DM \textit{forward} process produces a Markov chain $z_1, \dots, z_T$ by gradually adding Gaussian noise to $z_0$ based on a variance schedule $\beta_1, \dots, \beta_T$, that is:
\begin{small}
\begin{equation}
\label{eq:forward}
    q(z_t|z_{t-1}) = \mathcal{N}(z_t; \sqrt{1-\beta_t}z_{t-1}, \beta_t\mathbf{I})
\enspace,
\end{equation}
\end{small}

\noindent where variances $\beta_t$ are constants.
If $\beta_t$ are small, the posterior $q(z_{t-1}|z_{t})$ can be well approximated by diagonal Gaussian \cite{nichol2021glide,sohl2015deep}. Furthermore, when the $T$ of the chain is large enough, $z_T$ can be well approximated by standard Gaussian distribution $\mathcal{N}(\mathbf{0}, \mathbf{I})$. These suggest that the true posterior $q(z_{t-1}|z_{t})$ can be estimated by $p_\theta(z_{t-1}|z_t)$ defined as \cite{nichol2021improved}:
\begin{small}
\begin{equation}
\label{eq:reverse}
    p_\theta(z_{t-1}|z_t)=\mathcal{N}(z_{t-1}; \mu_\theta(z_t), \Sigma_\theta(z_t))
\enspace.
\end{equation}
\end{small}

\noindent The DM \textit{reverse} process (also known as \textit{sampling}) then generates samples $z_0\sim p_\theta(z_0)$ by initiating a Markov chain with Gaussian noise $z_T\sim \mathcal{N}(\mathbf{0}, \mathbf{I})$ and progressively decreasing noise in the chain of $z_{T-1}, z_{T-2}, \dots, z_0$ using the learnt $p_\theta(z_{t-1}|z_t)$. To learn $p_\theta(z_{t-1}|z_t)$, Gaussian noise $\epsilon$ is added to $z_0$ to generate samples $z_t\sim q(z_t|z_0)$, then a model $\epsilon_\theta$ is trained to predict $\epsilon$ using the following mean-squared error loss:
\begin{small}
\begin{equation}
\label{eq:dm}
    L_\text{DM}=\mathbb{E}_{t\sim \mathcal{U}(1, T), z_0\sim q(z_0), \epsilon\sim \mathcal{N}(\mathbf{0}, \mathbf{I})}[||\epsilon-\epsilon_\theta(z_t, t)||^2]
\enspace,
\end{equation}
\end{small}

\noindent where time step $t$ is uniformly sampled from $\{1, \dots, T\}$. Then $\mu_\theta(z_t)$ and $\Sigma_\theta(z_t)$ in Eq.~\ref{eq:reverse} can be derived from $\epsilon_\theta(z_t, t)$ to model $p_\theta(z_{t-1}|z_t)$ \cite{ho2020denoising,nichol2021improved}.
The denoising model $\epsilon_\theta$ is typically implemented using a time-conditioned U-Net \cite{ronneberger2015u} with residual blocks \cite{he2016deep} and self-attention layers \cite{vaswani2017attention}. Sinusoidal position embedding \cite{vaswani2017attention} is also usually used to specify the time step $t$ to $\epsilon_\theta$.


\subsection{HistoDiffusion}

\textbf{Model Architecture.}
Our proposed HistoDiffusion is built on Latent Diffusion Models (LDM)~\cite{rombach2022high}, which requires fewer computational resources without degradation in performance, compared to prior works \cite{dhariwal2021diffusion,ker2017deep,shen2017deep}. 
LDM first trains a latent autoencoder (LAE) \cite{kingma2013auto} to encode images as lower-dimensional latent representations and then learns a diffusion model (DM) for image synthesis by modeling the latent space of the trained LAE.
Particularly, the encoder $\mathcal{E}$ of the LAE encodes the input image $x\in\mathbb{R}^{H\times W\times 3}$ into a latent representation $z=\mathcal{E}(x)\in\mathbb{R}^{h\times w\times c}$ in a lower-dimensional latent space $\mathcal{Z}$. Here $H$ and $W$ are the height and width of image $x$, and $h$, $w$, and $c$ are the height, width, and channel of latent $z$, respectively.
The latent $z$ is then passed into the decoder $\mathcal{D}$ to reconstruct the image $\hat{x}=\mathcal{D}(z)$. Through this process, the compositional features from the image space $\mathcal{X}$ can be extracted to form the latent space $\mathcal{Z}$, and 
we then model the distribution of $\mathcal{Z}$ by learning a DM.
For the DM in LDM, both the forward and reverse sampling processes are performed in the latent space $\mathcal{Z}$ instead of the original image space $\mathcal{X}$. 

\noindent\textbf{Unconditional Large-scale Pre-training.}
To ensure the latent space $\mathcal{Z}$ can cover features of various data types, 
we first pre-train our proposed HistoDiffusion on large-scale unlabeled datasets.
Specifically, we gather unlabeled images from $M$ different sources to construct a large-scale set of datasets $\mathcal{S}=\{S_1, S_2, \dots, S_M\}$. 
We then train an LAE using the data from $\mathcal{S}$ with the following self-reconstruction loss to learn a powerful latent space $\mathcal{Z}$ that can describe diverse features:
\begin{small}
\begin{equation}
\label{eq:lae}
    L_\text{LAE} = \mathcal{L}_\text{rec}(\hat{x}, x) + \lambda_\text{KL}D_\text{KL}(q(z)||\mathcal{N}(\mathbf{0}, \mathbf{I}))\enspace,
\end{equation}
\end{small}

\noindent where $\mathcal{L}_\text{rec}$ is the loss measuring the difference between the output reconstructed image $\hat{x}$ and the input ground truth image $x$. Here we implement $\mathcal{L}_\text{rec}$ with a combination of a pixel-wise $L_1$ loss, a perceptual loss \cite{zhang2018unreasonable}, and a patch-base adversarial loss \cite{dosovitskiy2016generating,esser2021taming}. To avoid arbitrarily high-variance latent spaces, we also add a KL regularization term $D_\text{KL}$ \cite{kingma2013auto,rombach2022high} to constrain the variance of the latent space $\mathcal{Z}$ with a slight KL-penalty. 

After training the LAE, we fixed the trained encoder $\mathcal{E}$ and then train a DM with the loss $L_\text{DM}$ in Eq.~\ref{eq:dm} to model $\mathcal{E}$'s latent space $\mathcal{Z}$.
Here $z_0=\mathcal{E}(x)$ in Eq.~\ref{eq:dm}. Once the DM is trained, we can use denoising model $\epsilon_\theta$ in the DM reverse sampling process to synthesize a novel latent $\Tilde{z}_0\in\mathbb{R}^{h\times w\times c}$ and employ the trained decoder $\mathcal{D}$ to generate a new image $\Tilde{x}=\mathcal{D}(\Tilde{z}_0)$, which should satisfy the similar distribution as the data in $\mathcal{S}$.

\noindent\textbf{Conditional Small-scale Fine-tuning.}
Using the LAE and DM pretrained on $\mathcal{S}$, we can only generate the new image $\Tilde{x}$ following the similar distribution in $\mathcal{S}$. To generalize our HistoDiffusion to the small-scale labeled dataset $S^\prime$ collected from a different source (\textit{i.e.}, $S^\prime\not\subset\mathcal{S}$), we further fine-tune HistoDiffusion using the labeled data from $S^\prime$. Let $y$ be the label of image $x$ in $S^\prime$.
To minimize the training cost, we fix both the trained encoder $\mathcal{E}$ and trained DM model $\epsilon_\theta$ to keep latent space $\mathcal{Z}$ unchanged. Then we only fine-tune the decoder $\mathcal{D}$ using labeled data $(x, y)$ from $S^\prime$ with the following loss function: 
\begin{small}
\begin{equation}
\label{eq:dec}
    L_\mathcal{D} = \mathcal{L}_\text{rec}(\hat{x}, x) + \lambda_\text{CE}\mathcal{L}_\text{CE}(\varphi(\hat{x}), y)\enspace,
\end{equation}
\end{small}

\noindent where $\mathcal{L}_\text{rec}(\hat{x}, x)$ is the self-reconstruction loss between the output reconstructed image $\hat{x}=\mathcal{D}(\mathcal{E}(x))$ and the input ground truth image $x$. To enhance the correlation between the decoder output $\hat{x}$ and label $y$, we also add an auxiliary image classifier $\varphi$ trained with $(x, y)$ on the top of $\mathcal{D}$ and impose the cross-entropy classification loss $\mathcal{L}_\text{CE}$ when fine-tuning $\mathcal{D}$. 
$\lambda_\text{CE}$ is the balancing parameter. We annotate this fine-tuned decoder as $\mathcal{D}^\prime$ for differentiation.

\noindent\textbf{Classifier-guided Conditional Synthesis.}
To enable conditional image generation with our HistoDiffusion,
we further apply the classifier-guided diffusion sampling proposed in \cite{dhariwal2021diffusion,shi2023exploring,sohl2015deep,song2020score} using the labeled data $(x,y)$ from small-scale labeled dataset $S^\prime$. 
We first utilize the trained encoder $\mathcal{E}$ to encode the data $x$ from $S^\prime$ as latent $z_0$. Then we train a time-dependent latent classifier $\phi$ with paired $(z_t, y)$ using the following loss function:
\begin{small}
\begin{equation}
\label{eq:class}
L_\phi = \mathcal{L}_\text{CE}(\phi(z_t), y)\enspace,
\end{equation}
\end{small}

\noindent where $z_t\sim q(z_t|z_0)$ is the noisy version of $z_0$ at the time step $t$ during the DM forward process, and $\mathcal{L}_\text{CE}$ is the cross-entropy classification loss.
Based on the trained unconditional diffusion model $\epsilon_\theta$, and a classifier $\phi$ trained on noisy input $z_t$, 
we enable conditional diffusion sampling by perturbing the reverse-process mean with the gradient of the log probability $p_\phi(y|z_t)$ of a target class $y$ predicted by the classifier $\phi$ as follows:
\begin{small}
\begin{equation}
\label{eq:guidance}
    \hat{\mu}_\theta(z_t|y) = \mu_\theta(z_t) + g\cdot \Sigma_\theta(z_t)\nabla_{z_t}\log p_\phi(y|z_t) \enspace,
\end{equation}
\end{small}

\noindent where $g$ is the guidance scale. Then the DM reverse process in HistoDiffusion can finally generate a novel latent $\Tilde{z}_0$ satisfying the class condition $y$ through a Markov chain starting with a standard Gaussian noise $z_T\sim \mathcal{N}(\mathbf{0}, \mathbf{I})$ using $p_{\theta, \phi}(z_{t-1}|z_t, y)$ defined as follows:
\begin{small}
\begin{equation}
\label{eq:cond_reverse}
    p_{\theta, \phi}(z_{t-1}|z_t, y)=\mathcal{N}(z_{t-1}; \hat{\mu}_\theta(z_t|y), \Sigma_\theta(z_t))
\enspace. 
\end{equation}
\end{small}

\noindent The final image $\Tilde{x}$ of class $y$ can be generated by applying the fine-tuned decoder $\mathcal{D}^\prime$, \textit{i.e.}, $\Tilde{x}=\mathcal{D}^\prime(\Tilde{z_0})$.

\noindent\textbf{Selective Augmentation.}
To further improve the efficacy of synthetic augmentation, we follow~\cite{xue2021selective} to selectively add synthetic images to the original labeled training data based on centroid feature distance. The augmentation ratio is defined as the ratio between the selected synthetic images and the original training images. More results are demonstrated later in Table~\ref{tab:res}.

\section{Experiments}

\noindent\textbf{Datasets.}
We employ three public datasets of histopathology images during the large-scale pre-training procedure. The first one is the H\&E breast cancer dataset~\cite{claudio2021pathologygan}, containing 312,320 patches extracted from the hematoxylin \& eosin (H\&E) stained human breast cancer tissue micro-array (TMA) images~\cite{marinelli2007stanford}. Each patch has a resolution of $224\times224$. The second dataset is PanNuke~\cite{gamper2019pannuke}, a pan-cancer histology dataset for nuclei instance segmentation and classification. The PanNuke dataset includes 7,901 patches of 19 types of H\&E stained tissues obtained from multiple data sources, and each patch has a unified size of $256\times256$ pixels. The third dataset is TCGA-BRCA-A2/E2~\cite{van2021deepmed}, a subset derived from the TCGA-BRCA breast cancer histology dataset~\cite{cancer2012comprehensive}. The subset consists of 482,958 patches with a resolution of $256\times256$. Overall, there are 803,179 patches used for pre-training. 
As for fine-tuning and evaluation, we employ the NCT-CRC-HE-100K dataset that contains 100,000 patches from H\&E stained histological images of human colorectal cancer (CRC) and normal tissue. The patches have been divided into 9 classes: Adipose (ADI), background (BACK), debris (DEB), lymphocytes (LYM), mucus (MUC), smooth muscle (MUS), normal colon mucosa (NORM), cancer-associated stroma (STR), colorectal adenocarcinoma epithelium (TUM). The resolution of each patch is $224\times224$. 


To replicate a scenario where only a small annotated dataset is available for training, we have opted to utilize a subset of 5,000 (5\%) samples for fine-tuning. This subset has been carefully selected through an even sampling without replacement from each tissue type present in the train set. It is worth noting that the labels for these samples have been kept, which allows the fine-tuning process to be guided by labeled data, leading to better predictions on the specific task or domain being trained. By ensuring that the fine-tuning process is representative of the entire dataset through even sampling from each tissue type, we can eliminate bias towards any particular tissue type.
We evaluate the fine-tuned model on the official test set. The related data use declaration and acknowledgment can be found in our supplementary materials.

\noindent\textbf{Evaluation Metrics.}
We employ Fréchet Inception Distance (FID) score~\cite{heusel2017gans} to assess the image quality of the synthetic samples. We further compute the accuracy, F1-score, sensitivity, and specificity of the downstream classifiers to evaluate the performance gain from different augmentation methods. 





\begin{figure}[t]
\centering
\includegraphics[width=0.98\textwidth]{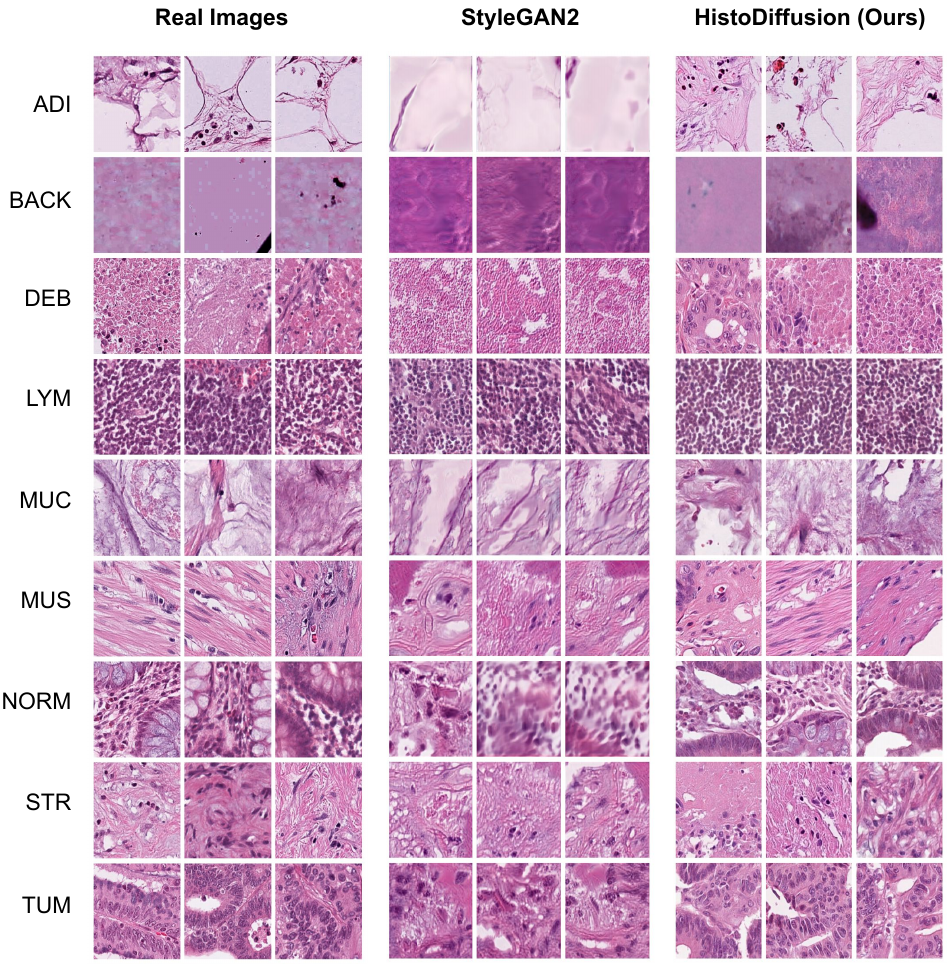}
\caption{Comparison of real images from training subset, synthesized images generated by StyleGAN2~\cite{karras2020analyzing} and our proposed HistoDiffusion (zoom in for clear observation). Qualitatively, our synthesized results contain more realistic and diagnosable patterns than results synthesized from StyleGAN2.
} 
\label{fig:syn_comparison}
\vspace{-7mm}
\end{figure}

\noindent\textbf{Model Implementation.}
All the patches are resized to $256\times256\times3$ before being passed into the models. Our implementation of HistoDiffusion basically follows the LDM-4~\cite{rombach2022high} architecture, where the input is downsampled by a factor of 4, resulting in a latent representation with dimensions of $64\times64\times3$. We use 1000 timesteps ($T=1000$) for the training of diffusion model and sample with classifier-free guidance scale $g=1.0$ and $200$ DDIM steps.
The latent classifier $\phi$ is constructed using the encoder architecture of the LAE and an additional attention pooling layer~\cite{radford2021learning} added before the output layer. 
 
We use the same architecture for the auxiliary image classifier $\varphi$. 
For downstream evaluation, we implement the classifier using the ViT-B/16 architecture \cite{dosovitskiy2020image} in all experiments to ensure fair comparisons. The default hyper-parameter settings provided in their officially released codebases are followed. 



\begin{table*}[tbp]
\caption{Quantitative comparison results of synthetic image quality and augmented classification. ``Random'' refers to directly augmenting the training dataset with synthesized images without any image selections while ``selective'' indicates applying selective module \cite{xue2021selective} to filter out low-quality images. The number ($X\%$) suggests that the number of the synthesized images is $X\%$ of the original training set.}
\vspace{-4mm}
\begin{center}
\resizebox{\textwidth}{!}{%
\begin{tabular}{r | >{\centering}p{1.7cm}  || >{\centering}p{1.7cm}  | >{\centering}p{1.7cm} | >{\centering}p{2cm} | c }
\hline
{} &      FID$\downarrow$ &   Accuracy$\uparrow$ &              F1 Score$\uparrow$ &      Sensitivity$\uparrow$ &      Specificity$\uparrow$ \\
\hline
Baseline (5\% real images)       
& / 
&  0.855
&  0.850
&   0.855
&   0.983\\
\hhline{=|=|=|=|=|=}

StyleGAN2~\cite{karras2020analyzing}   \\
\hline
\enspace + random 50\% 
& 5.714
& 0.860
& 0.856
& 0.860
& 0.980
\\
\hline
\enspace + selective~\cite{xue2021selective} 50\%     
& 5.088
& 0.868
& 0.861
& 0.867
& 0.978
 \\
\hline

100\%  
&  5.927
& 0.879
& 0.876
& 0.879
& 0.982
 \\
\hline

200\%  
& 7.550
& 0.895
& 0.888
& 0.895
& 0.983
 \\
\hline

300\%  
& 10.643
& 0.898
& 0.896
& 0.898
& 0.987
\\
\hline

\hhline{=|=|=|=|=|=}

HistoDiffusion (Ours)     \\
\hline
\enspace + random 50\% 
& 4.921
& 0.870
& 0.869
& 0.870
& 0.982
\\
\hline
\enspace + selective~\cite{xue2021selective} 50\%      
&  4.544
&  0.891
&  0.888
&  0.891
&  0.983
 \\
\hline

100\%  
&  \textbf{3.874}
&  0.903
& 0.902
& 0.903
& 0.991
 \\
\hline

200\%  
&{4.583}
&\textbf{0.919}
&\textbf{0.916}
&\textbf{0.919}
&\textbf{0.992}
\\
\hline

300\%  
& 8.326
& 0.910
& 0.912
& 0.910
& 0.988
\\
\hline
\end{tabular}
}
\end{center}
\label{tab:res}
\vspace{-5mm}
\end{table*}

\noindent\textbf{Comparsion to state-of-the-art.}
We compare our proposed HistoDiffusion with the current state-of-the-art cGAN-based method~\cite{xue2021selective}. We employ StyleGAN2~\cite{karras2020analyzing} as the backbone generative model for cGAN-based synthesis. To ensure a fair comparison, all images synthesized by StyleGAN2 and HistoDiffusion model are further selected based on feature centroid distances~\cite{xue2021selective}. More implementation details of our proposed HistoDiffusion, StyleGAN2, and baseline classifier can also be found in our supplementary materials. 

\noindent\textbf{Result Analysis.}
As shown in Table~\ref{tab:res}, under the same synthetic augmentation setting, HistoDiffusion shows better FID scores and outperforms the state-of-the-art cGAN model StyleGAN2 in all classification metrics. A qualitative comparison between synthetic images by HistoDiffusion and StyleGAN2 can be found in Fig.~\ref{fig:syn_comparison}, where HistoDiffusion consistently generates more realistic images matching the given class conditions than SytleGAN2, especially for classes ADI and BACK.

When augmenting the training dataset with different numbers of images synthesized from HistoDiffusion and StyleGAN2, one can observe that when increasing the ratio of synthesized data to 100\%, the FID score of StyleGAN2 increases quickly and can become even worse than the one without using image selection strategy. In contrast, HistoDiffusion can keep synthesizing high-quality images until the augmentation ratio reaches 300\%. Regarding classification performance improvement of the baseline classifier, the accuracy and F1 score of using HistoDiffusion augmentation are increased by up to \textbf{6.4}\% and \textbf{6.6}\%, respectively.
Even when not using the image selection module to filter out the low-quality results (\textit{i.e.}, +random 50\%), our HistoDiffusion can still improve the accuracy by 1.5\%. The robustness and effectiveness of HistoDiffusion can be attributed to the unconditional large-scale pre-training, our specially-designed conditional fine-tuning, and classifier-guided generation, among others.

\section{Conclusions}
In this study, we have introduced a novel synthetic augmentation technique, termed HistoDiffusion, to enhance the performance of medical image recognition systems. HistoDiffusion leverages multiple unlabeled datasets for large-scale, unconditional pre-training, while employing a labeled dataset for small-scale conditional fine-tuning. Experiment results on a histopathology image dataset excluded from the pre-training demonstrate that given limited labels, HistoDiffusion with image selection remarkably enhances the classification performance of the baseline model, and can potentially handle any future incoming small dataset for augmented training using the same pre-trained model. 

\bibliographystyle{splncs04}
\bibliography{mybib}

\clearpage
\setcounter{figure}{0}
\setcounter{table}{0}

\thispagestyle{empty}
\section*{\centering {\LARGE Supplementary Materials}}
\begin{table}[h]
\caption{
Quantitative ablation study results of synthetic image quality and augmented classification performance from HistoDiffusion without conditional small-scale fine-tuning and HistoDiffusion trained with only a small-scaled dataset (5\% real images). Compared with the final HistoDiffusion, the model without conditional fine-tuning generates images that drastically differ from real images, resulting in high FID scores, whilst the model trained without any pre-training leads to worse classification performance. For the best model trained using 200\% synthetic images, the accuracy and F1 score are still \textbf{2.4\%} and \textbf{2.2\%} lower than the final model, respectively.
}
\resizebox{\linewidth}{!}{%
\begin{tabular}{r | >{\centering}p{1.7cm}  || >{\centering}p{1.7cm}  | >{\centering}p{1.7cm} | >{\centering}p{2cm} | c }
\hline
{} &      FID$\downarrow$ &   Accuracy$\uparrow$ &              F1 Score$\uparrow$ &      Sensitivity$\uparrow$ &      Specificity$\uparrow$ \\

\hline
\hline

HistoDiffusion \\
(w/o conditional small-scale fine-tuning)   \\
\hline
\enspace + random 50\% 
& \textbf{18.254}
& 0.865
& 0.867
& 0.865
& 0.986

\\
\hline
\enspace + selective 50\%     
& 19.887
& 0.875
& 0.870
& 0.875
& 0.982

 \\
\hline

100\%  
& 18.418
& 0.880
& 0.879
& 0.880
& 0.985

 \\
\hline

200\%  
& 19.229
& 0.897
& 0.892
& 0.897
& \textbf{0.989}

 \\
\hline

300\%  
& 21.682
& \textbf{0.905} 
& \textbf{0.908}
& \textbf{0.905}
& 0.988

\\
\hline
\hline

HistoDiffusion \\
(w/o unconditional large-scale pre-training)     \\
\hline
\enspace + random 50\% 
& 5.687
& 0.870
& 0.867
& 0.870
& 0.984

\\
\hline
\enspace + selective 50\%      
& 5.256
& 0.884
& 0.882
& 0.884
& 0.986

 \\
\hline

100\%  
& 4.387
& 0.890
& 0.891
& 0.890
& \textbf{0.988}

 \\
\hline

200\%  
& \textbf{3.618}
& \textbf{0.895}
& \textbf{0.894}
& \textbf{0.895}
& \textbf{0.988}

\\
\hline

300\%  
& 4.446
& 0.892
& 0.887
& 0.892
& 0.983

\\
\hline
\end{tabular}
}
\label{tab:supp_res}
\end{table}

\thispagestyle{empty}
\begin{table}[ht]
\caption{
Implementation details of each modules in HistoDiffusion, baseline classifier, and the baseline StyleGAN2.
All models are trained on NVIDIA Quadro RTX 6000 GPUs with 24GB RAM.}
\begin{center}
\resizebox{\linewidth}{!}{%
\begin{tabular}{l | >{\centering}p{1.8cm}  | >{\centering}p{1.7cm}  | >{\centering}p{1.7cm} | >{\centering}p{1.7cm} | >{\centering}p{1.7cm}|  >{\centering}p{1.7cm} | c }
\hline
{} 
& Latent \\Autoencoder
& Diffusion Models
& Latent Classifier
& Auxiliary Classifier
& Fine-tuned Decoder
& Baseline Classifier
& StyleGAN2
\\
\hline
Batch size    
& 6
& 4
& 32
& 64
& 4
& 128
& 64
\\
\hline
Training epochs
& 10
& 10
& 100
& 100
& 50
& 200
& 100
 \\
\hline
Learning rate   
& 4.5e-6
& 5.0e-5
& 5.0e-5
& 5.0e-5
& 5.0e-5
& 1.0e-4
& 2.0e-4
\\
\hline
Optimizer
& Adam
& AdamW
& AdamW
& Adam
& Adam
& AdamW
& Adam
\\
\hline

Loss weights $\lambda$
& 1.0e-6 ($\lambda_\text{KL}$)
& -
& -
& -
& 1.0 ($\lambda_\text{CE}$)
& -
& -
\\
\hline

Guidance scale $g$
& -
& -
& 1.0
& -
& -
& -
& -
 \\
\hline

Noise schedule
& -
& Linear
& -
& -
& -
& -
& -
 \\
\hline

\end{tabular}
}
\end{center}
\vspace{-5mm}
\end{table}

\thispagestyle{empty}

\begin{figure}[h]
\centering
\includegraphics[width=0.98\linewidth]{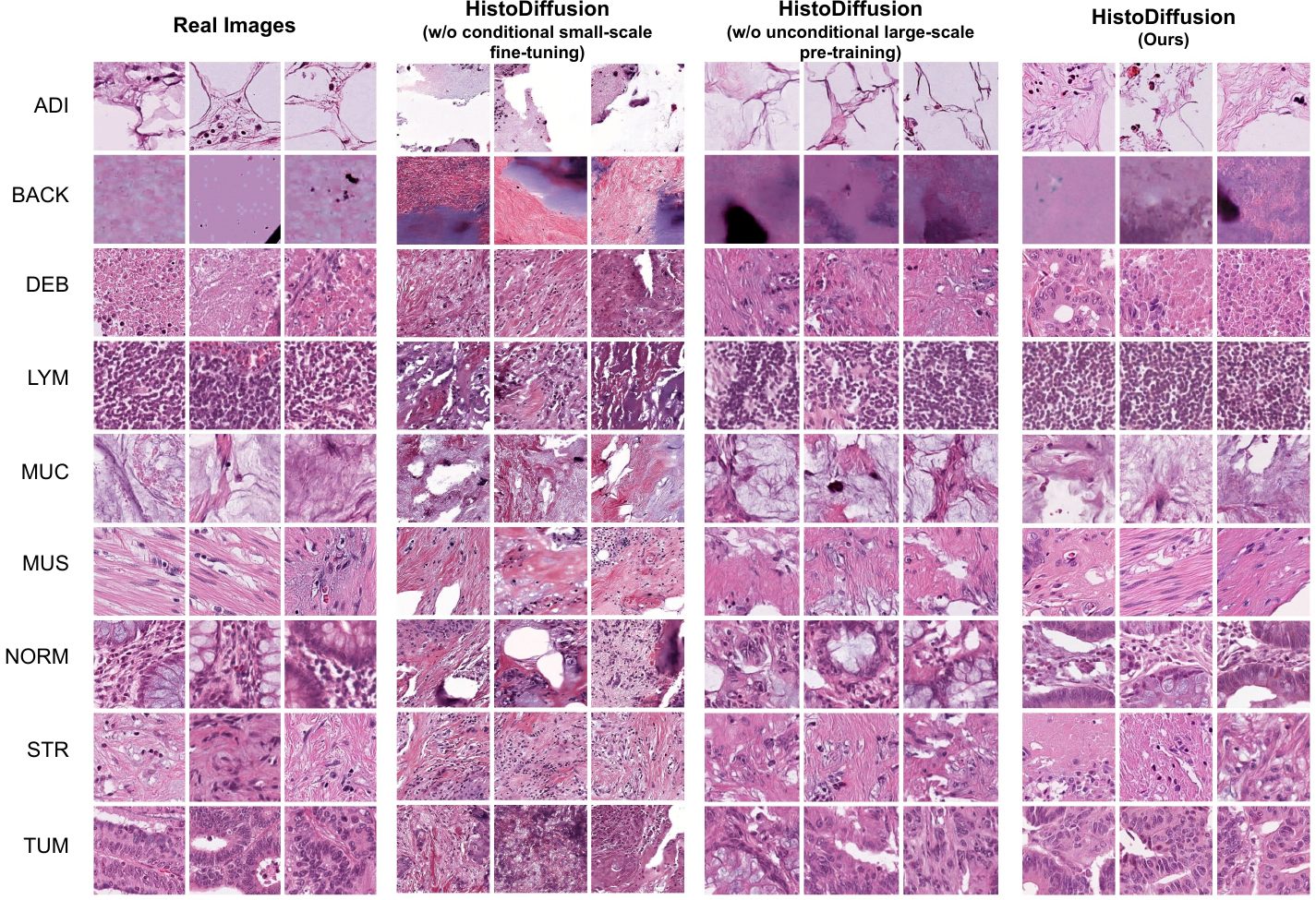}
\caption{
Qualitative visualization of synthetic images generated by our proposed HistoDiffusion,  HistoDiffusion without conditional small-scale fine-tuning, and HistoDiffusion trained using only a small-scaled dataset of 5\% real images (\textit{i.e.,} without using unconditional large-scale pre-training). Compared with our final HistoDiffusion, [HistoDiffusion without conditional fine-tuning] sometimes fails to generate images that contain correct corresponding characteristics implied by the class condition (\textit{e.g.,} NORM and LYM) while [HistoDiffusion without unconditional large-scale pre-training] struggles to synthesize diverse results (\textit{e.g.,} BACK and MUS). Zoom in for better details.
}
\end{figure}

\end{document}